\documentclass{isprs} 
\usepackage{setspace}
\usepackage{geometry} 
\usepackage{epstopdf}
\usepackage[labelsep=period]{caption}  
\usepackage[british]{babel} 
\usepackage{todonotes}
\usepackage{multirow}
\usepackage{booktabs}
\usepackage{float}
\usepackage{placeins}
\usepackage{url}

\usepackage{tikz}
\usepackage{pgfplots}

\usepackage{array}
\newcolumntype{L}[1]{>{\raggedright\let\newline\\\arraybackslash\hspace{0pt}}m{#1}}
\newcolumntype{C}[1]{>{\centering\let\newline\\\arraybackslash\hspace{0pt}}p{#1}}
\newcolumntype{R}[1]{>{\raggedleft\let\newline\\\arraybackslash\hspace{0pt}}m{#1}}

\usepackage{xcolor}
\definecolor{forest}{HTML}{009900}
\definecolor{shrubland}{HTML}{c6b044}
\definecolor{savanna}{HTML}{fbff13}
\definecolor{grassland}{HTML}{b6ff05}
\definecolor{wetlands}{HTML}{27ff87}
\definecolor{croplands}{HTML}{c24f44}
\definecolor{urban}{HTML}{a5a5a5}
\definecolor{snow}{HTML}{69fff8}
\definecolor{barren}{HTML}{f9ffa4}
\definecolor{water}{HTML}{1c0dff}

\makeatletter
\newenvironment{customlegend}[1][]{%
    \begingroup
    \let\addlegendimage=\pgfplots@addlegendimage
    \let\addlegendentry=\pgfplots@addlegendentry
    \pgfplots@init@cleared@structures
    \pgfplotsset{#1}%
}{
    \pgfplots@createlegend
    \endgroup
}
\makeatother

\begin{document}

\title{Weakly Supervised Semantic Segmentation of Satellite Images for Land Cover Mapping -- Challenges and Opportunities}

\author{
  M. Schmitt\textsuperscript{1}, J. Prexl\textsuperscript{1}, P. Ebel\textsuperscript{1}, L. Liebel\textsuperscript{2}, X.X. Zhu\textsuperscript{1,3}
}

\address{
	\textsuperscript{1 }Signal Processing in Earth Observation, Technical University of Munich, Munich, Germany - (m.schmitt,jonathan.prexl)@tum.de\\
 	\textsuperscript{2 }Chair of Remote Sensing Technology, Technical University of Munich, Munich, Germany - lukas.liebel@tum.de\\
	\textsuperscript{3 }Remote Sensing Technology Institute, German Aerospace Center (DLR), Oberpfaffenhofen, Germany - xiaoxiang.zhu@dlr.de
}



\abstract{Fully automatic large-scale land cover mapping belongs to the core challenges addressed by the remote sensing community. Usually, the basis of this task is formed by (supervised) machine learning models. However, in spite of recent growth in the availability of satellite observations, accurate training data remains comparably scarce. On the other hand, numerous global land cover products exist and can be accessed often free-of-charge. Unfortunately, these maps are typically of a much lower resolution than modern day satellite imagery. Besides, they always come with a significant amount of noise, as they cannot be considered ground truth, but are products of previous (semi-)automatic prediction tasks. Therefore, this paper seeks to make a case for the application of weakly supervised learning strategies to get the most out of available data sources and achieve progress in high-resolution large-scale land cover mapping. Challenges and opportunities are discussed based on the SEN12MS dataset, for which also some baseline results are shown. These baselines indicate that there is still a lot of potential for dedicated approaches designed to deal with remote sensing-specific forms of weak supervision.
}


\maketitle


\sloppy

\section{Introduction}\label{sec:Intro}
The problem of automatic land cover mapping from remote sensing imagery is traditionally cast as a (supervised) machine learning task, especially when applied to large study areas \cite{Cihlar2000}. However, while the amount of available satellite data keeps on growing, training labels remain rare, because of the difficulty to create reliable land cover annotations that can be referred to as ``ground truth''. On the other hand, manifold large-scale land cover datasets already exist, all of which are the result of (semi-)automated processes themselves. This introduces \emph{weakly supervised learning} as a promising strategy to train well-generalizing models on available data -- even if the labels come with a significant error bar or at comparably low resolutions. 

In this paper, we discuss the problem of weakly supervised learning of models for land cover prediction from satellite data. For this purpose, we focus on the freely available global imagery provided by the Sentinel-1 and Sentinel-2 missions of the European Copernicus program \cite{Torres2012,Drusch2012} and a simplified version of the land cover classification scheme of the International Geosphere-Biosphere Programme (IGBP) \cite{Loveland1997}, which is reflected by the \emph{SEN12MS} dataset \cite{Schmitt2019} and the 2020 IEEE-GRSS Data Fusion Contest (DFC2020) \cite{Yokoya2020}. Besides a description of the challenge and how \emph{SEN12MS} and \emph{DFC2020} are addressing it, baseline results using off-the-shelf deep learning models are provided to highlight the importance of dedicated research.  
 
\section{Weakly Supervised Learning}
In his excellent review, \cite{Zhou2018} defines weakly supervised learning as an umbrella term addressing the attempt to construct predictive models from three types of weak supervision:
\begin{itemize}
    \setlength{\itemsep}{-1pt}
    \item \emph{Incomplete supervision}\newline
    In this case, a small amount of labeled data, which is insufficient to train a good model and abundant unlabeled data are available. 
    \item \emph{Inexact supervision}\\
    In this case, some supervision information is given, but it is not as exact as necessary. An example of this is land cover labels, which have a lower resolution than the satellite observations that shall be processed. 
    \item \emph{Inaccurate supervision}\\
    In this case, annotations cannot be considered as ground-truth; i.e., at least some of the labels are erroneous. 
\end{itemize}\vspace{-8pt}
In the context of this paper, weakly supervised learning is restricted to the cases of inexact and inaccurate supervision, which can also be seen as different forms of label noise. In contrast to that, incomplete supervision is seen as a different case, which is addressed by \emph{semi-supervised learning} \cite{Zhu2009}, which is not covered here. As shown in the following sections, dealing with different forms of noisy samples has become a well-addressed field in machine learning and should receive quite some attention by remote sensing researchers as well.

\subsection{Machine Learning with Noisy Samples}
Weakly supervised learning in the above-defined sense, i.e. learning from inexact and inaccurate samples, has become a sub-field in machine learning research that has been drawing a significant amount of interest. While there are some studies, which indicate that deep neural networks are relatively robust to label noise \cite{Rolnick2018}, many researchers investigate approaches to deal with this challenge based on insights from robust statistics and dedicated mathematical modelling. Thus, popular solutions in this context are either the formulation of robust loss functions, e.g. \cite{Ghosh2017}, the iterative improvement of training data via bootstrapping \cite{Reed2015}, or the addition of dedicated noise layers to the neural network \cite{Sukhbaatar2015}. As summarized in \cite{Frenay2014}, it can be stated that numerous possible coping strategies exist. 

\subsection{Relevance for the Remote Sensing of Land Cover}
Remote sensing has long been a primary source of big data \cite{Chi2016}, with the numbers of available observations and measurements of our planet continuously on the rise. Driven by this development, deep learning has drawn significant attention from the research community \cite{Zhu2017}. However, as highlighted by \cite{Reichstein2019}, the lack of dedicated large training or benchmark datasets still remains one of the grand challenges in the creation of operational models for real-world applications. On the other hand, past efforts of remote sensing scientists and practitioners have led to the production of numerous large-scale -- or even global -- land cover maps. As nicely summarized by \cite{Grekousis2015}, the resolutions of those maps typically range from $30$m to $1{,}000$m per pixel with overall accuracies between $64\%$ and $88\%$. In other words, plenty of noisy training labels are potentially available free of charge! Inspired by the generic techniques for machine learning from noisy samples described in the previous section, one would think that weakly supervised learning of land cover prediction models using these available datasets as training input would have become a major theme in modern day remote sensing research. Interestingly, however, the literature dedicated to this challenge is still rather scarce. While most papers addressing weak supervision in a remote sensing context deal with object detection, e.g. \cite{Zhang2015,Kellenberger2019}, the few papers addressing weakly supervised semantic segmentation usually rely on sparse or even only image-level annotations, e.g. \cite{Fu2018,Wang2020}, instead of coarse and/or noisy labels available in a dense manner. A quite notable exception is the work by \cite{Robinson2019}, who fused low-resolution and high-resolution labels in order to produce a high-resolution land cover map of the contiguous United States. Their approach is based on what they called \emph{super-resolution loss} in an earlier contribution \cite{Malkin2019}, which allows to predict high-resolution land cover from low-resolution labels by modeling the expected distribution of high-resolution land cover and using its distance to the predicted distribution as an additional loss term.

Using the \emph{SEN12MS} dataset, which combines noisy land cover labels with a resolution of $500$m with Sentinel-1 SAR and Sentinel-2 optical data, as an example, this paper seeks to provide a basis for further explorations of weakly supervised semantic segmentation of satellite images for land cover prediction.

\begin{figure*}[h!]
    \centering
    \includegraphics[width=0.7\linewidth]{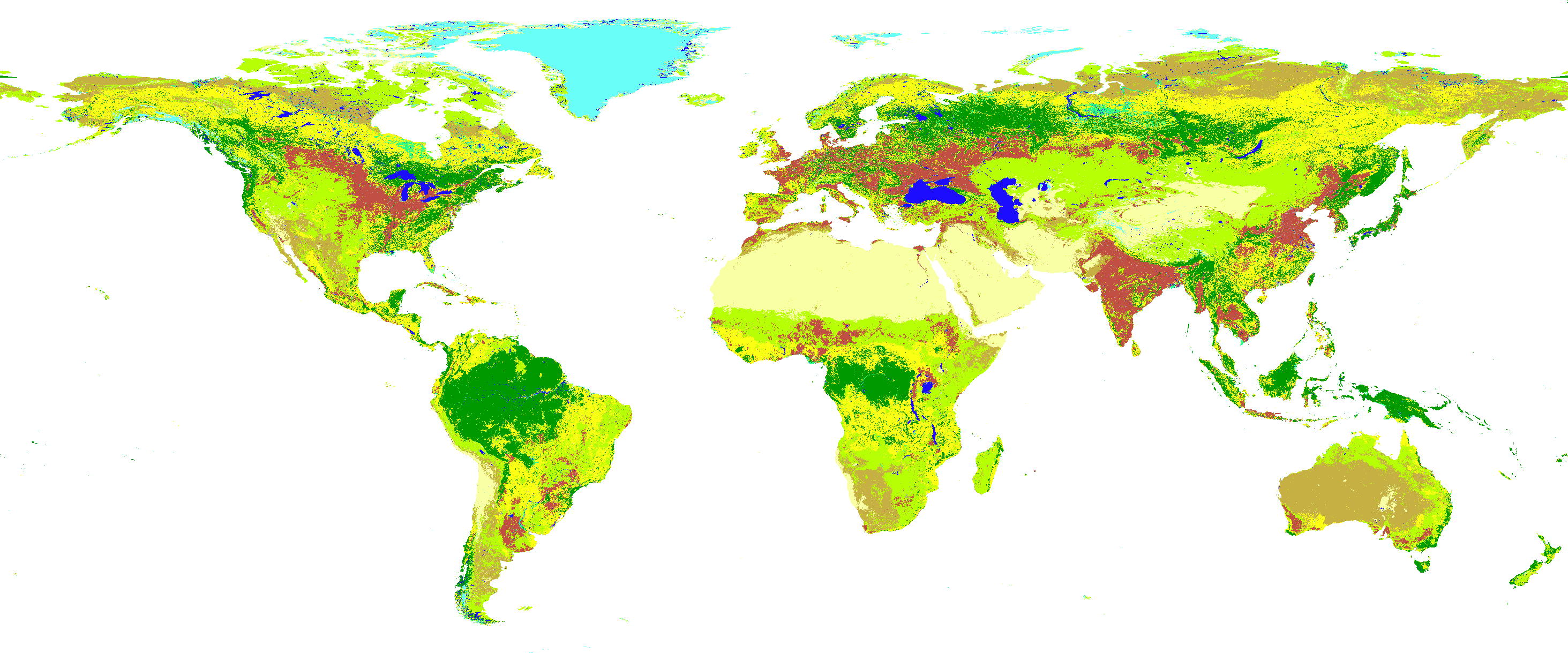}
    \begin{tikzpicture}
    \pgfplotsset{
    legend style={cells={anchor=west}, draw=none,column sep=1ex,
    nodes={scale=0.7, transform shape}}
    }
    \begin{customlegend}[legend columns=10]
    \addlegendimage{forest, only marks, mark=square*}
    \addlegendentry{Forest}
    \addlegendimage{shrubland, only marks, mark=square*}
    \addlegendentry{Shrubland}
    \addlegendimage{savanna, only marks, mark=square*}
    \addlegendentry{Savanna}
    \addlegendimage{grassland, only marks, mark=square*}
    \addlegendentry{Grassland}
    \addlegendimage{wetlands, only marks, mark=square*}
    \addlegendentry{Wetland}
    \addlegendimage{croplands, only marks, mark=square*}
    \addlegendentry{Cropland}
    \addlegendimage{urban, only marks, mark=square*}
    \addlegendentry{Urban}
    \addlegendimage{snow, only marks, mark=square*}
    \addlegendentry{Snow}
    \addlegendimage{barren, only marks, mark=square*}
    \addlegendentry{Barren}
    \addlegendimage{water, only marks, mark=square*}
    \addlegendentry{Water}
    \end{customlegend}
    \end{tikzpicture}
    \caption{The MODIS-derived world map at a resolution of $500$m following the simplified IGBP scheme. While larger areas of the \emph{Savanna} class (in yellow color) are found in South America, Africa and Australia as expected, there are also vast areas of \emph{Savanna} in Canada, Scandinavia, and Siberia -- regions in which savannas in the geographical sense of the term usually do not exist.}
    \label{fig:IGBP_worldMap}
\end{figure*}

\section{Weakly Supervised Learning for Land Cover Mapping with SEN12MS}\label{sec:main}
\begin{table*}
\centering\footnotesize
    \begin{tabular}{L{1.5cm} L{4.5cm} L{1.5cm} L{3cm} L{1.5cm}}
     \toprule
        IGBP Class Number & IGBP Class Name & Simplified Class Number & Simplified Class Name & Color\\
        \cmidrule{1-1}  \cmidrule(lr){2-2} \cmidrule(lr){3-3} \cmidrule(lr){4-4} \cmidrule{5-5} 
        
        1  & Evergreen Needleleaf Forest & \multirow{5}{=}{1} & \multirow{5}{=}{Forest} & \multirow{5}{=}{\textcolor{forest}{009900}}\\
        2  & Evergreen Broadleaf Forest & &    & \\ 
        3  & Deciduous Needleleaf Forest & & & \\
        4  & Deciduous Broadleaf Forest & & & \\
        5 & Mixed Forest & & &\\
        \addlinespace
        
        6 & Closed Shrublands & \multirow{2}{=}{2} & \multirow{2}{=}{Shrubland} & \multirow{2}{=}{\textcolor{shrubland}{c6b044}}\\
        7 & Open Shrublands & & & \\
        \addlinespace
        
        8 & Woody Savannas & \multirow{2}{=}{3} & \multirow{2}{=}{Savanna} & \multirow{2}{=}{\textcolor{savanna}{fbff13}}\\
        9 & Savanna & & & \\\addlinespace\addlinespace
        
        10 & Grasslands & 4 & Grassland & \textcolor{grassland}{b6ff05}\\\addlinespace\addlinespace
        
        11 & Permanent Wetlands & 5 & Wetlands & \textcolor{wetlands}{27ff87}\\\addlinespace\addlinespace
        
        12 & Croplands &  \multirow{2}{=}{6} & \multirow{2}{=}{Croplands} & \multirow{2}{=}{\textcolor{croplands}{c24f44}}\\
        14 & Cropland / Natural Vegetation Mosaics & & & \\\addlinespace\addlinespace
        
        13 & Urban and Built-up Lands & 7 & Urban/Built-up & \textcolor{urban}{a5a5a5}\\\addlinespace\addlinespace
        
        15 & Permanent Snow and Ice & 8 & Snow/Ice & \textcolor{snow}{69fff8}\\\addlinespace\addlinespace
        
        16 & Barren & 9 & Barren & \textcolor{barren}{f9ffa4}\\\addlinespace\addlinespace
        
        17 & Water Bodies & 10 & Water & \textcolor{water}{1c0dff}\\
    \bottomrule
    \end{tabular}
    \caption{The simplified IGBP land cover classification scheme.}
    \label{tab:IGBP_simple}
\end{table*}

The \emph{SEN12MS} dataset \cite{Schmitt2019} was published in 2019 as the largest curated dataset dedicated to deep learning in remote sensing at that time. It consists of $180{,}662$ patch triplets sampled over all meteorological seasons and all inhabited continents in order to represent a global distribution. Every triplet consists of a dual-polarimetric Sentinel-1 SAR image, a multi-spectral Sentinel-2 image tensor, and four different land cover maps following different internationally established classification schemes. In the frame of the 2020 IEEE-GRSS Data Fusion Contest (DFC2020), the organizers defined the weakly supervised training of globally applicable land cover prediction models as the contest goal \cite{Yokoya2020}.

\subsection{The Simplified IGBP Land Cover Classification Scheme}\label{sec:IGBP_simple}
For the DFC2020 the IGBP classification scheme, which originally is comprised of 17 classes \cite{Loveland1997}, was aggregated to 10 less fine-grained classes (see Tab.~\ref{tab:IGBP_simple}). This \emph{simplified IGBP scheme} is similar to the classification scheme adopted by the authors of the FROM-GLC10 dataset \cite{Gong2019}, which constitutes the first global land cover map with a resolution of $10$m (at an overall validation accuracy of about $73\%$). Both schemes differ in only one class: While the simplified IGBP scheme contains a \emph{Savanna} class, the FROM-GLC10 scheme contains a \emph{Tundra} class. However, both classes are restricted to certain geographical regions: 
According to the Encyclopedia Britannica, a savanna \textit{``is characterized by an open tree canopy (i.e., scattered trees) above a continuous tall grass understory (the vegetation layer between the forest canopy and the ground)''}. Mostly found \textit{``in Africa, South America, Australia, India, the Myanmar (Burma)-Thailand region in Asia, and Madagascar''}, savannas thus are a land cover type, which can not be found around the globe, but only in specific geographical regions. Above that, they are also not suitable for classical pixel-based classification approaches, since at a resolution of $10$m no \emph{Savanna} pixels exist -- one will either find pixels containing trees (i.e. the \emph{Forest} class in simplified IGBP terms), or grass (i.e. \emph{Grassland}). At a resolution of $500$m, however, the mixing of the spectral responses of sparse trees and grass understory can well lead to a distinct spectral \emph{Savanna} profile. It has to be noted that this is different for approaches that take spatial context into account as do, for example, convolutional neural networks -- as long as their receptive field is large enough.

As can be seen in Fig.~\ref{fig:IGBP_worldMap}, in the MODIS-derived IGBP land cover map, which constitutes the basis of the \emph{SEN12MS} land cover annotations, the \emph{Savanna} class is way more widely spread than one would expect based on the above-mentioned definition, even outside those regions where savannas actually exist, which should be considered as a form of systematic label noise. For generic solutions to global land cover mapping, it will thus be advisable to adapt suitable strategies that either ignore training pixels with \emph{Savanna} label, or that allow a transformation of \emph{Savanna} into classes such as \emph{Grassland}, or \emph{Forest}, which are applicable in all regions of the world. Since there is certainly no one-to-one mapping between \emph{Savanna} and the alternative classes, statistical strategies such as, e.g., the one proposed by \cite{Malkin2019} are in need.

\subsection{SEN12MS}\label{sec:SEN12ms_stats}
\begin{figure}[htb]
    \centering
    \includegraphics[width = .65\linewidth]{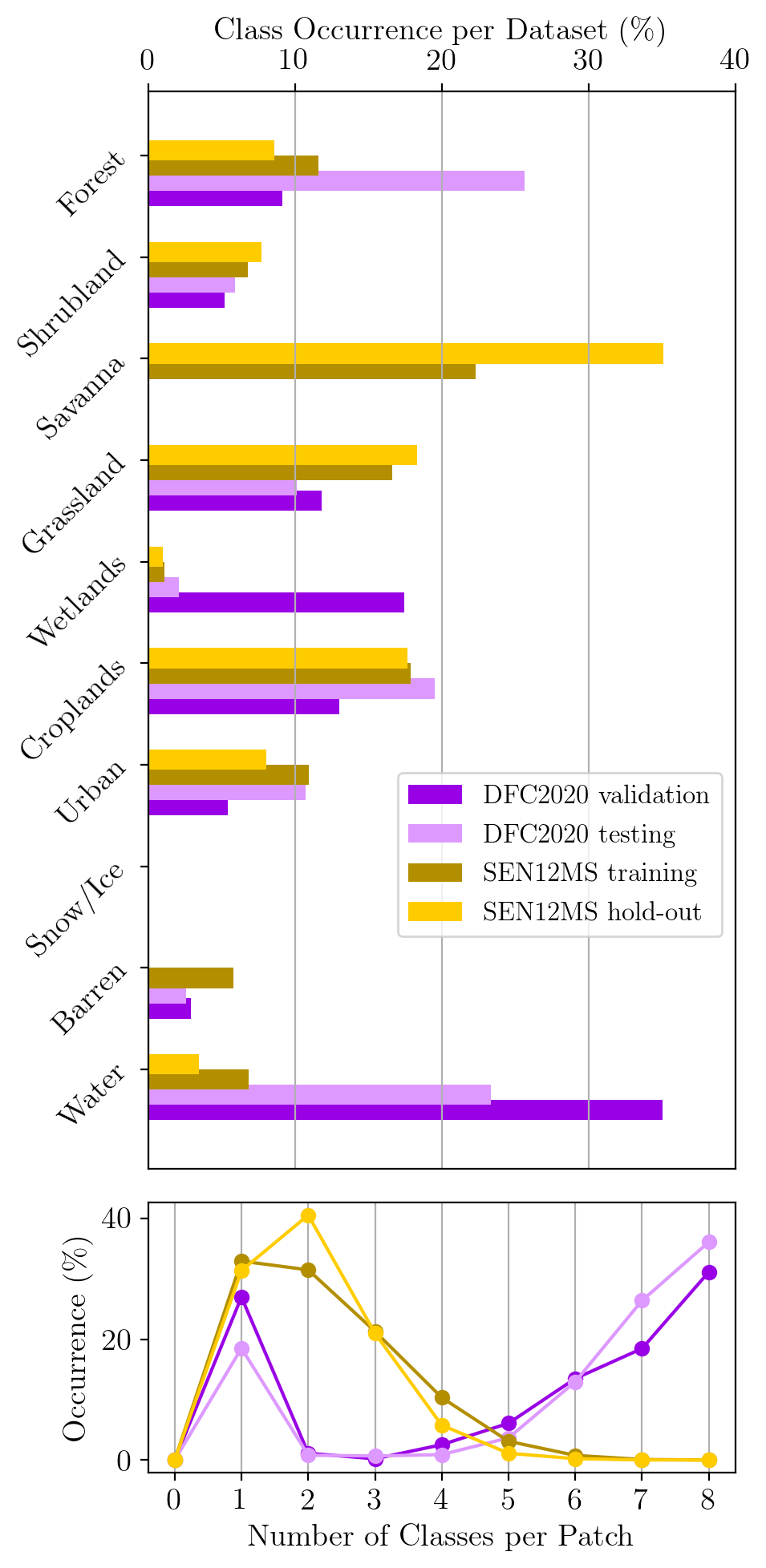}
    \caption{The distribution for the different land cover classes (top) as well as the
    number of different classes per $256\times 256$-pixels image patch (bottom).}
    \label{fig:class_occurence}
\end{figure}

The distribution of the pixels contained in the \emph{SEN12MS} dataset over the 10 classes of the simplified IGBP scheme is shown in Fig.~\ref{fig:class_occurence}. While the distribution is relatively balanced in terms of the classes \emph{Forest}, \emph{Grassland}, \emph{Croplands}, and \emph{Urban}, the classes \emph{Shrubland}, \emph{Barren}, and \emph{Water} are slightly less frequent. The major outliers are the classes \emph{Wetlands} and \emph{Snow/Ice}, which hardly exist, and the largest class \emph{Savanna}, which accounts for almost a quarter of all pixels in the dataset.

The reason for this imbalancing are multifaceted:
Firstly, wetlands, for example, are simply relatively rare in reality. Apart from that, water areas were purposefully undersampled because of the simplicity to map them, whereas urban areas were purposefully oversampled because of their heterogeneity and also their importance to geographical science. As discussed in Section~\ref{sec:IGBP_simple}, the \emph{Savanna} class is over-represented because the global, MODIS-derived IGBP land cover map, which constitutes the basis of the \emph{SEN12MS} land cover annotations, contains much more savanna areas than one would expect. This has to be considered when using the dataset as basis for the development of land cover-oriented semantic segmentation models.

\subsection{DFC2020}
For the IEEE-GRSS 2020 Data Fusion Contest, a high-resolution (GSD: $10$m) dataset for validation and testing was generated in a semi-manual manner, following the simplified IGBP scheme as well. While the \emph{DFC2020 validaton} and \emph{test} labels and all relevant meta-information will only officially be published after the end of the contest in April 2020, the class distributions of the data are already shown in Fig.~\ref{fig:class_occurence} for sake of comparison with \emph{SEN12MS}. It can be seen that the distributions are fairly similar. The most important exception is the complete absence of the \emph{Savanna} class. This is due to the fact that the \emph{DFC2020} maps were created in a semi-manual manner and on a pixel-level basis for regions of interest outside the typical savanna regions. Besides the absence of the \emph{Savanna} class, also no \emph{Snow/Ice} pixels exist in the \emph{DFC2020} data.

As can be seen from Fig.~\ref{fig:class_occurence}, another interesting difference between the \emph{SEN12MS} and the \emph{DFC2020} datasets is the fact that the high-resolution \emph{DFC2020} patches either contain a single class (e.g. in homogeneous \emph{Forest} or \emph{Water} areas) -- or more than five classes, whereas the low-resolution MODIS-derived labels of \emph{SEN12MS} mostly contain one to three classes. This is a clear hint towards the significant resolution difference.

\subsection{Predicting High-Resolution Land Cover from Low-Resolution Labels}
With the availability of \emph{SEN12MS} for training and \emph{DFC2020} for validation and/or testing, a wide range of possibilities for weakly supervised training of high-resolution land cover prediction models opens up. To make full use of them it is crucial to have a common understanding of the data structures. A summary is given in Tab.~\ref{tab:subsets}.
\begin{table}[h]
    \centering
    \footnotesize
    \begin{tabular}{L{1.2cm} R{1cm} L{4.5cm}}
     \toprule
     Dataset &  Size & Comment\\
       \cmidrule{1-1}  \cmidrule(lr){2-2} \cmidrule{3-3} 
        \emph{SEN12MS training} & $162{,}556$ & subset of SEN12MS dedicated to training\\
        \addlinespace
        \emph{SEN12MS hold-out} & $18{,}106$ & hold-out set with low-res labels; similar spatial and temporal distribution as the overall dataset; used for validation or testing\\
        \addlinespace
        \emph{DFC2020 validation} & $986$ & used for testing in the first phase of DFC2020, and for validation in the second phase\\
        \addlinespace
        \emph{DFC2020 testing} & $5{,}128$ & used for testing in the second phase of DFC2020\\
         \bottomrule
    \end{tabular}
    \caption{The different sub-datasets that can be built from the \emph{SEN12MS} and \emph{DFC2020} data.}
    \label{tab:subsets}
\end{table}
While it is perfectly possible to keep all $180{,}662$ patches of \emph{SEN12MS} in a single dataset purely used for training, and all $6{,}114$ patches of \emph{DFC2020} in another dataset purely used for testing, we suggest to make use of the splits proposed in this paper in future work  to ensure comparability between achieved results in a benchmarking sense. The list of hold-out scenes for \emph{SEN12MS} can be found in the \emph{SEN12MS} support repository at \url{https://github.com/schmitt-muc/SEN12MS}, and the \emph{DFC2020} data is provided in separate validation and test packages at  \url{https://ieee-dataport.org/competitions/2020-ieee-grss-data-fusion-contest}.

\section{Baseline Results}\label{sec:Baseline}
To provide a first intuition about what is possible when using the \emph{SEN12MS} and \emph{DFC2020} datasets for weakly supervised learning, first results are collected in this section. They shall also serve as examples for future benchmarking purposes. While land cover maps are traditionally assessed via the \emph{overall accuracy} (OA) measure, we propose to use the less optimistic \emph{average accuracy} (AA) for comparison, as it gives less weight to large classes, which are rather simple to classify, e.g. \emph{Forest} and \emph{Water}. It is important to note that in this paper, AA refers to average \emph{producer's} accuracy, which is highly correlated to the often-used mean intersection over union (mIoU) metric.

To implement the considerations about the difficult \emph{Savanna} class described in Section~\ref{sec:main}, during training of all machine learning models, \emph{Savanna} pixels were not used. 

With respect to the satellite input data, the following pre-processing was applied: The Sentinel-1 backscatter values were clipped and normalized to the interval $[-25, 0]$, before rescaling to $[0,1]$. In a similar manner, we clipped the intensity values of the Sentinel-2 top-of-atmosphere observations to $[0, 10^4]$, corresponding to a maximum of $100\%$ surface reflectance before rescaling as well. It is important to note that we made only use of the 10 surface-related Sentinel-2 bands (i.e. the bands with an original resolution of $10$m and $20$m), while the atmosphere-related bands (with an original resolution of $60$m) B1, B9 and B10 were not used.

\subsection{Low-Resolution vs. High-Resolution Labels}
\begin{figure}[h!]
    \centering\scriptsize
    \rotatebox{90}{~~~~~~~~~~~~~~~~~~~~~~~~~~~~~~~~~~~~~~~~MODIS-derived labels}
    \includegraphics[width=.8\linewidth]{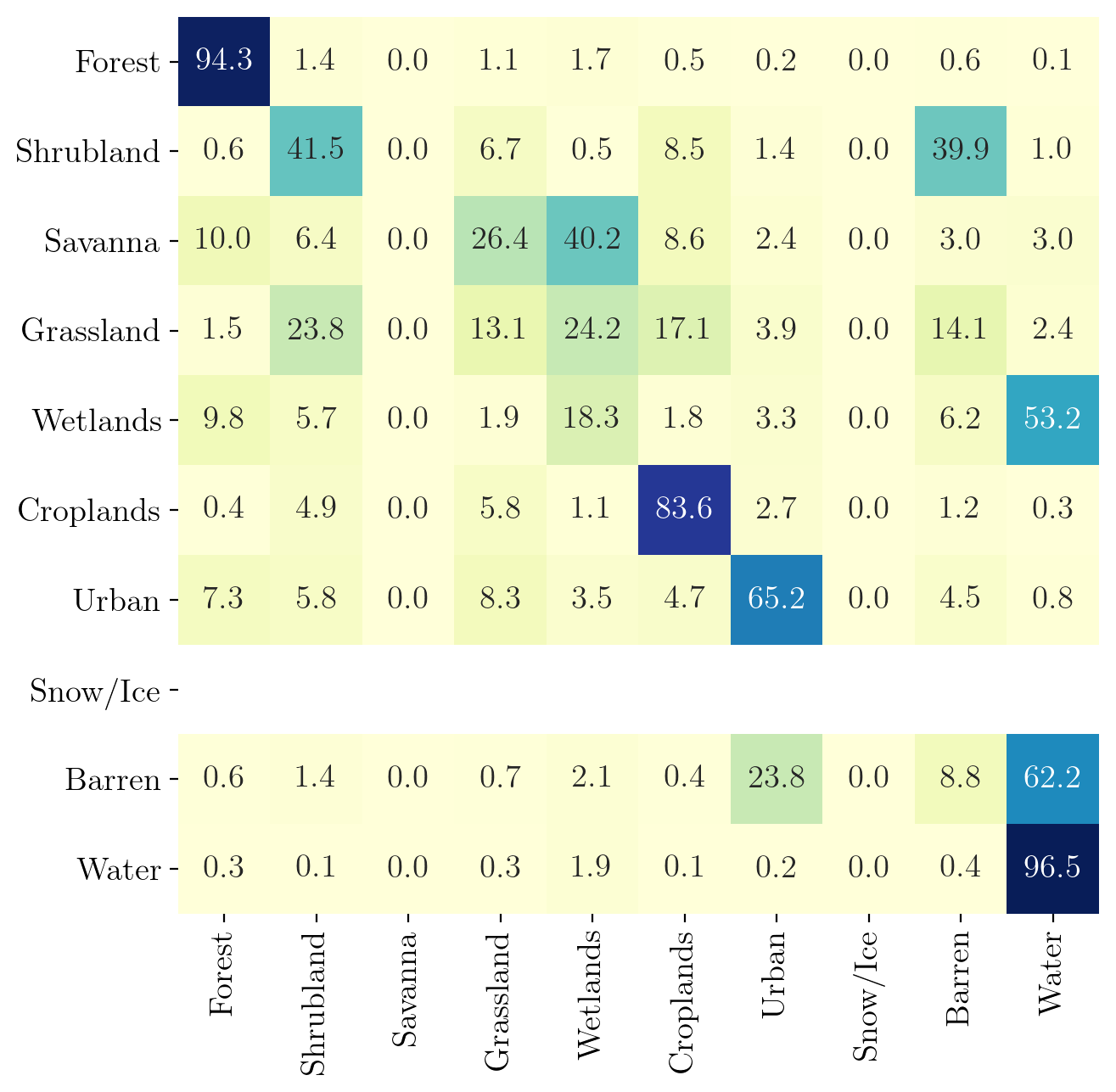}\\ ~~~~~~~~~~~~~~~~~~\emph{DFC2020} labels
    \caption{Class transition matrix from low-resolution, MODIS-derived labels to high-resolution \emph{DFC2020} labels.}
    \label{fig:transMat_LR_HR}
\end{figure}

As a sanity check and the lower end of what is possible, the low-resolution MODIS-derived labels can simply be tested against the high-resolution \emph{DFC2020 validation} set. The results are shown in the leftmost column of Tab.~\ref{tab:results_lowResHighRes}.
\begin{table*}[h]
    \centering
    \footnotesize
    \begin{tabular}{L{1.3cm} R{1cm} R{1cm} R{1cm} R{1cm} R{1cm} R{1cm} R{1cm} R{1cm} R{1cm}}
     \toprule
     Class &  LR-HR & DLv3 S2~only & DLv3 S1+S2 & Unet S2~only & Unet S1+S2 & $k$-means S2~only & $k$-means S1+S2& RF S2~only & RF S1+S2\\
       \cmidrule(r){1-1}  \cmidrule(lr){2-2} \cmidrule(lr){3-3} \cmidrule(lr){4-4} \cmidrule(lr){5-5} \cmidrule(lr){6-6} \cmidrule(lr){7-7} \cmidrule(lr){8-8} \cmidrule(lr){9-9} \cmidrule(l){10-10} 
Forest 		& 51.6\%	& 71.4\%  	& 61.2\% 	& 67.3\% 	& 55.4\%	& 2.4\% 	& 1.7\% 	& 77.1\% 	& 76.9\%	\\
Shrubland 	& 7.7\% 	& 2.3\%  	& 3.8\%  	& 0.0\% 	& 3.7\%		& 7.7\% 	& 5.9\% 	& 0.0\% 	& 0.0\% 	\\
Savanna 	& -- 		& -- 		& -- 		& -- 		& --		& -- 		& -- 		& -- 		& --		\\
Grassland 	& 6.7\% 	& 64.4\% 	& 48.2\% 	& 76.7\% 	& 77.2\%	& 11.2\% 	& 12.5\% 	& 90.3\% 	& 90.5\%	\\
Wetlands 	& 0.6\% 	& 2.4\% 	& 3.8\% 	& 3.7\% 	& 3.2\%		& 2.2\% 	& 0.3\% 	& 4.1\% 	& 4.0\% 	\\
Croplands 	& 64.4\% 	& 53.3\% 	& 61.9\% 	& 65.7\% 	& 50.7\%	& 42.1\% 	& 13.4\% 	& 42.1\% 	& 39.6\% 	\\
Urban 		& 71.5\% 	& 71.0\% 	& 62.8\% 	& 80.9\% 	& 73.1\%	& 0.0\% 	& 0.0\% 	& 0.0\% 	& 0.0\% 	\\
Snow/Ice 	& -- 		& -- 		& -- 		& -- 		& --		& -- 		& -- 		& -- 		& --		\\
Barren 		& 0.3\% 	& 0.2\% 	& 1.0\% 	& 0.6\% 	& 0.8\%		& 54.4\% 	& 6.2\% 	& 0.0\% 	& 0.0\%		\\
Water 		& 95.1\% 	& 88.9\% 	& 95.8\% 	& 89.4\% 	& 92.7\%	& 55.8\% 	& 68.9\% 	& 25.4\% 	& 34.5\%	\\
\textbf{Average}& \textbf{37.2\%} & \textbf{44.2\%} & \textbf{42.3\%} & \textbf{48.1\%} & \textbf{44.6\%} & \textbf{22.0\%} & \textbf{13.6\%} & \textbf{29.9\%} & \textbf{30.7\%}\\
         \bottomrule
    \end{tabular}
    \caption{Class-wise and average accuracies achieved on the \emph{DFC2020 validation} dataset for different benchmarks. \emph{S2 only} indicates that only Sentinel-2 data have been used for the prediction, whereas \emph{S1+S2} indicates the case of Sentinel-1/Sentinel-2 data fusion. \emph{LR-HR} indicates the baseline check of evaluating the MODIS-derived low-resolution labels against the high-resolution DFC2020 reference labels.}
    \label{tab:results_lowResHighRes}
\end{table*}
While frequent and easy-to-determine classes such as \emph{Forest}, \emph{Urban}, and \emph{Water} show relatively good agreement between the low-resolution labels and the high-resolution reference, less frequent classes, which are harder to identify (e.g. \emph{Shrubland}, \emph{Barren}, and \emph{Wetlands}) cause the average accuracy to drop to a mere $37.2\%$. On the other hand, it seems a bit surprising that the \emph{Croplands} class also shows a satisfying agreement, although empty fields could certainly be confused with \emph{Barren} or crops growing up with \emph{Grassland}. On the opposite, the \emph{Grassland} class shows an unexpectedly bad accuracy, which is mainly due to a confusion with \emph{Shrubland} or \emph{Wetlands} pixels in the high-resolution reference. More details can be seen from the class transition matrix shown in Fig.~\ref{fig:transMat_LR_HR}, which depicts the likelihood of a class in the high-resolution \emph{DFC2020} data given a class in the low-resolution MODIS-derived land cover map. The good agreement of \emph{Forest}, \emph{Croplands}, \emph{Urban}, and \emph{Water} are confirmed, while the confusion-prone classes \emph{Shrubland}, \emph{Grassland}, \emph{Wetlands} and \emph{Barren} can be further interpreted. While the transition of a \emph{Wetlands} pixel into a \emph{Water} pixel can be relatively comprehensible, the transition of \emph{Barren} pixels into \emph{Water} pixels can be considered a relevant potential source for label noise.

\subsection{Off-the-Shelf Models for Semantic Segmentation}\label{sec:off-the-shelf}
To provide a baseline for future developments, Tab.~\ref{tab:results_lowResHighRes} also contains the results for off-the-shelf models for semantic segmentation. All of them were trained on the \emph{SEN12MS training} subset, validated with the \emph{SEN12MS hold-out} subset, and tested on the \emph{DFC2020 validation} set, which was already officially available during the writing of this paper.  We implemented the ignoring of the \emph{Savanna} pixels during training using a masked-cross-entropy loss.

\begin{itemize}
    \setlength{\itemsep}{-1pt}
    \item \emph{DeepLabv3+ (DLv3)}\newline
Achieving top-ranking results on various semantic segmentation benchmarks, DeepLabv3+ \cite{Chen2018} represents a state-of-the-art semantic segmentation architecture and was, thus, used for our baseline experiments. 
Our implementation used a ResNet-101 backbone with ImageNet pre-trained weights as an initialization. In order for results to be comparable, we fixed the hyperparameters for training. Training was conducted for ten epochs.
\item \emph{Unet}\newline
In addition to \emph{DLv3} we further applied a \emph{Unet} type architecture \cite{ronneberger2015u} to the segmentation task.
We adopted the last layer to contain nine segmentation maps and masked the loss function to ignore the neglected tenth class.
The model contains $\approx 31$ million (random initialized) parameters, and is therefore significantly larger then the  \emph{DLv3}. Another important difference is the utilization of long skip connections in the \emph{Unet} architecture, which is expected to have a positive influence on preserving fine spatial details. 
\end{itemize}
Our Pytorch-based implementations of the two baseline networks are available at \url{https://github.com/lukasliebel/dfc2020_baseline}.

\begin{figure}[h]
    \centering\scriptsize
    \rotatebox{90}{~~~~~~~~~~~~~~~~~~~~~~~~~~~~~~~~~~~~~~~~\emph{DFC2020} reference data}
    \includegraphics[width=0.8\linewidth]{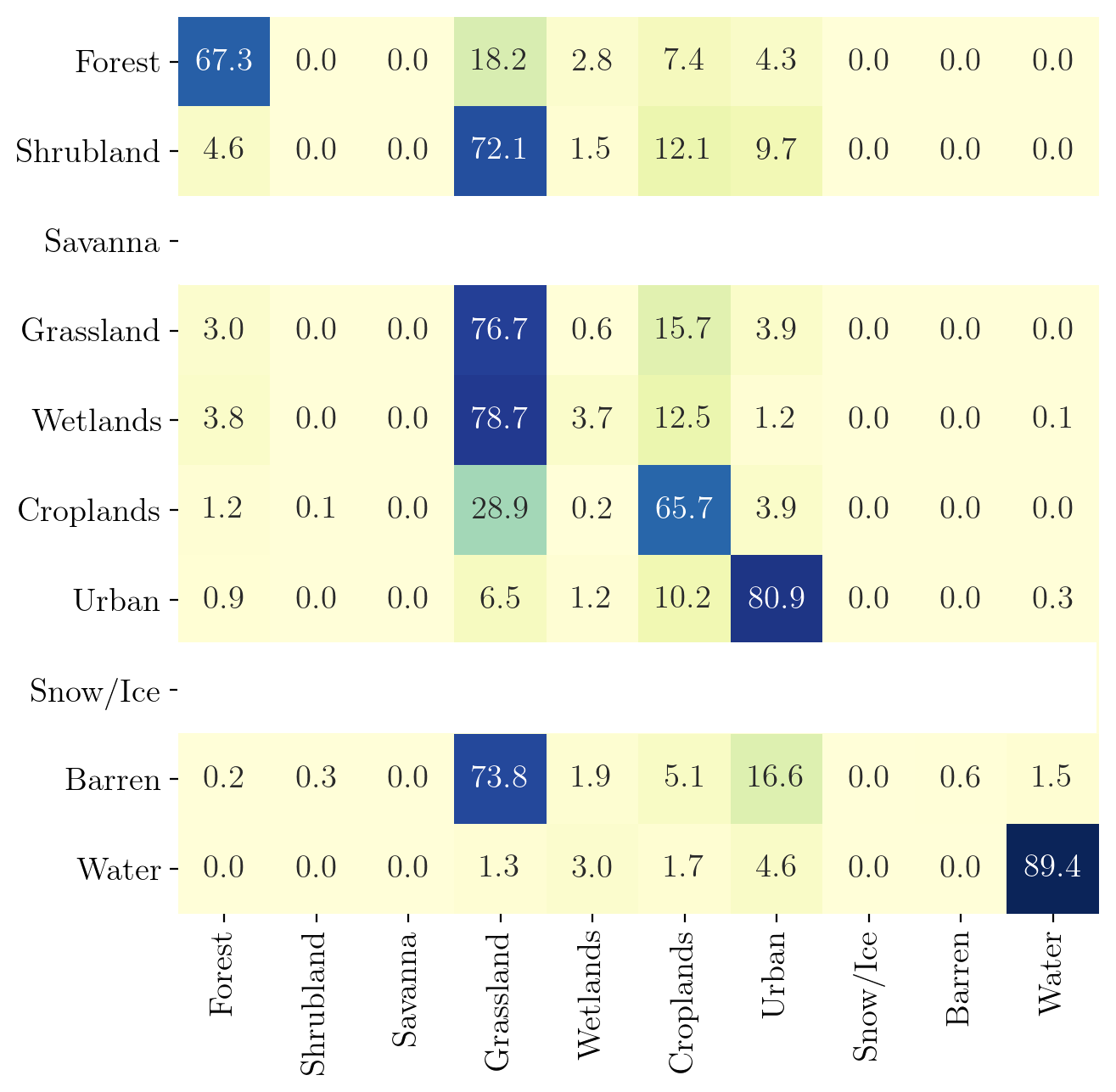}\\
    ~~~~~~~~~~~~~~~~~~~Unet-based predictions
    \caption{Confusion matrix of the Unet model using only Sentinel-2 data as input.}
    \label{fig:cm_unet}
\end{figure}

\begin{figure}
    \centering
    \includegraphics[width=.8\linewidth]{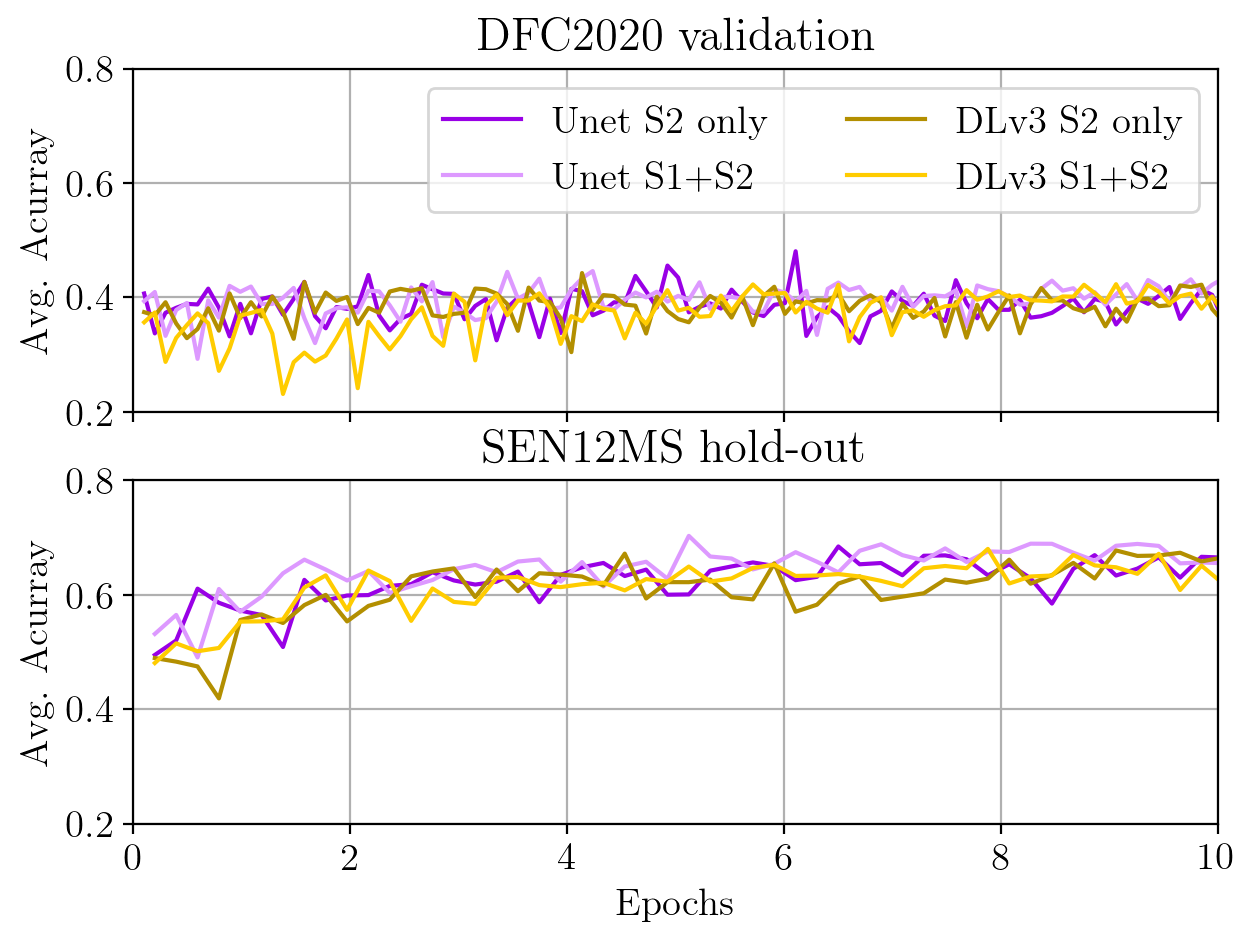}
    \caption{Comparison of the Average Accuracy metric achieved on both the \emph{DFC2020 validation} as well as the \emph{SEN12MS hold-out} set over the training process for the different deep semantic segmentation networks described in section~\ref{sec:off-the-shelf}. }
    \label{fig:dlv3_val_curves}
\end{figure}

Figure~\ref{fig:dlv3_val_curves} compares how the validation accuracy on the \emph{SEN12MS hold-out set} and the test accuracy on the  \emph{DFC2020 validation} set
change over time for the different models. Since the training is
carried out in on the low-resolution MODIS-derived land cover labels
without any specific adaptations to cope with the  situation  of  weak 
supervision, the slightly positive trend of the validation accuracy is not mirrored by the test accuracy -- the evolution of the
networks remains unstable. In order to fill Tab.~\ref{tab:results_lowResHighRes}, we select the
checkpoint with the best test accuracy for evaluation. This should be
seen as the upper bound of what is achievable with off-the-shelf
semantic segmentation networks and does not allow a judgment between the models.
The confusion matrix achieved by the best deep semantic segmentation network (i.e. the Unet relying on only Sentinel-2) is shown in Fig.~\ref{fig:cm_unet}.

The results show that even off-the-shelf semantic segmentation models produce results that are significantly better than the low-resolution reference. The most notable improvement is observed for the \emph{Grassland} class, with also \emph{Forest}, \emph{Croplands}, \emph{Urban}, and \emph{Water} performing reasonably well. On the downside, \emph{Wetlands} and \emph{Barren} are not really mapped well, whereas \emph{Shrubland} becomes even worse than in the low-resolution input. The main source of confusion is the \emph{Grassland} class, which collects most predictions from \emph{Shrubland}, \emph{Wetlands} and \emph{Barren} classes. With erroneous \emph{Grassland} predictions also affecting the \emph{Forest}, \emph{Shrubland} and \emph{Croplands} reference classes, this land cover type will need more attention in future model designs.

Figure~\ref{fig:pixelwise_segm} provides a visual impression of the mapping quality, taking a prediction example after the first epoch. While the off-the-shelf deep semantic segmentation models are able to recover the general scene structure, fine details get completely lost. 

\subsection{Shallow Learning Baselines}\label{sec:shallow}
To provide further baselines for the problem at hand, we trained two shallow classifiers -- one unsupervised, one supervised -- on a subset of the \emph{SEN12MS training} set. Both classifiers were trained on just $2{,}500$ patches uniformly sub-sampled from the full dataset. Due to the pixel-wise classification approach this amounts to an effective training data set size of about $164$ million individual observations. The sub-sampling is required due to the computational complexity of parts of the training, such as the Kuhn-Munkres algorithm having a run time of $\mathcal{O}(n^3)$.

\begin{figure}
\centering
\begin{tabular}{rcc}
 \rotatebox{90}{Sentinel-2 RGB} &
 \includegraphics[width=0.38\linewidth]{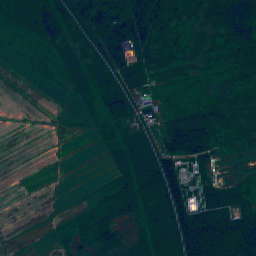}    &
 \includegraphics[width=0.38\linewidth]{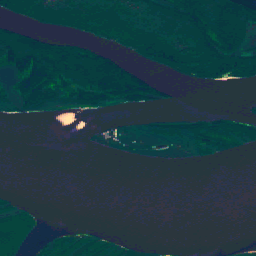}    \\
 \rotatebox{90}{\emph{DFC2020} reference} &
 \includegraphics[width=0.38\linewidth]{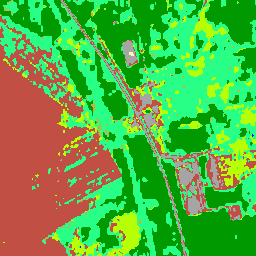} &
 \includegraphics[width=0.38\linewidth]{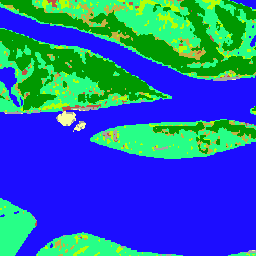}\\
 \rotatebox{90}{MODIS LC}&
 \includegraphics[width=0.38\linewidth]{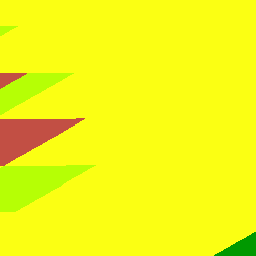}    &
 \includegraphics[width=0.38\linewidth]{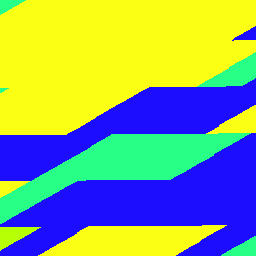}    \\
 \rotatebox{90}{DLv3} &
  \includegraphics[width=0.38\linewidth]{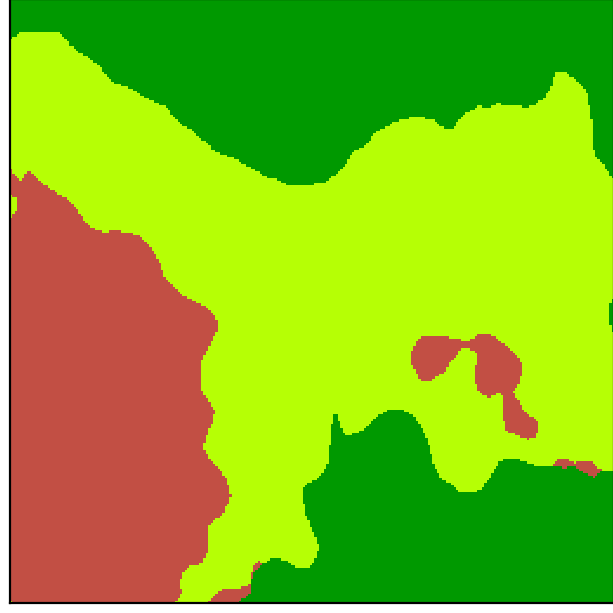} &
 \includegraphics[width=0.38\linewidth]{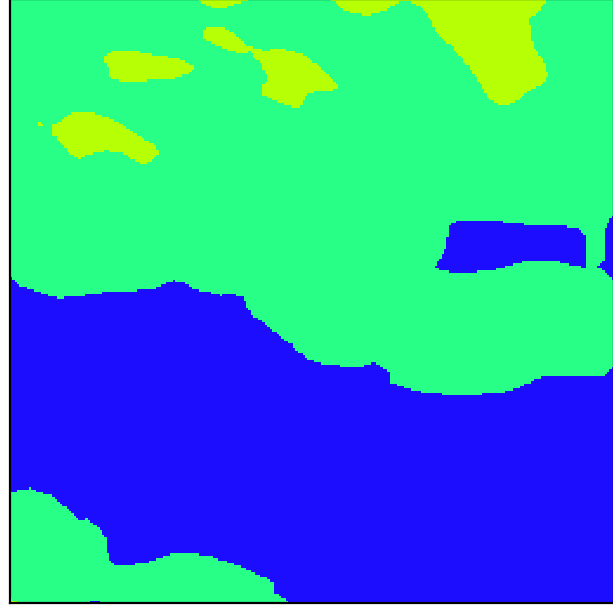}\\ 
 \rotatebox{90}{Unet} &
   \includegraphics[width=0.38\linewidth]{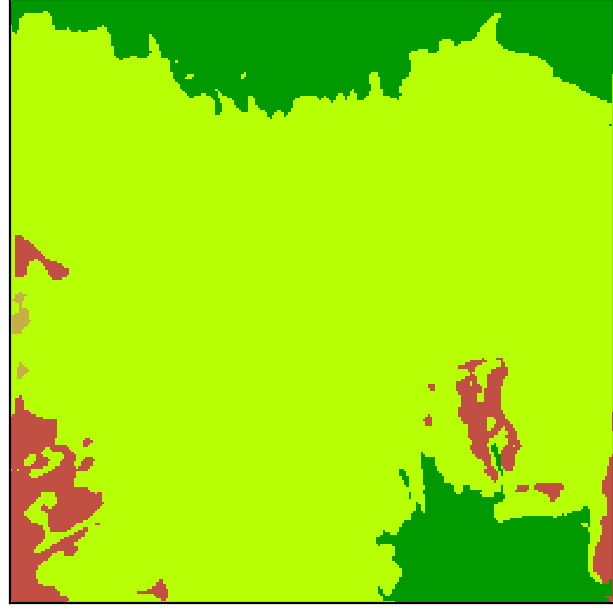}  & 
 \includegraphics[width=0.38\linewidth]{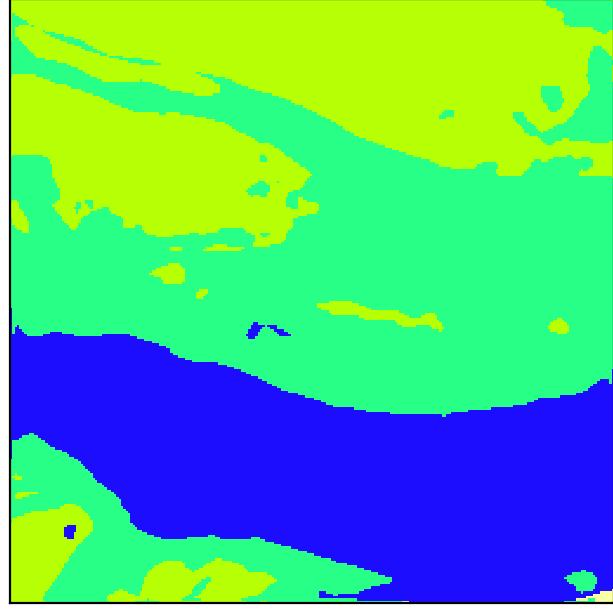}  \\
 \rotatebox{90}{$k$-means} &
 \includegraphics[width=0.38\linewidth]{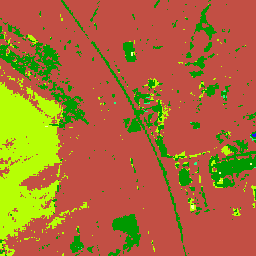}    &
 \includegraphics[width=0.38\linewidth]{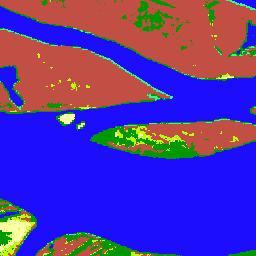}    \\
 \rotatebox{90}{RF} &
 \includegraphics[width=0.38\linewidth]{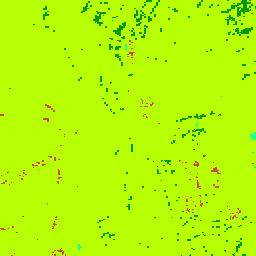}  & 
 \includegraphics[width=0.38\linewidth]{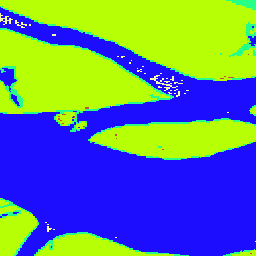}  \\
\end{tabular}
\caption{Land cover mapping results achieved with the baseline models for two example patches. Note that for Unet and DLv3 the models based on Sentinel-1 and Sentinel-2 data fusion and achieved after the first training epoch were used to predict the maps shown in this figure, as they provided a visual upper bound.}
\label{fig:pixelwise_segm}
\end{figure}

Details of the two setups are described in the following. Both used just the above-described 12 channels of Sentinel-1 and Sentinel-2 as pixel-wise input features.\vspace{-6pt}
\begin{itemize}
\setlength{\itemsep}{-1pt}
\item \emph{$k$-means clustering}\newline
$k=8$ clusters, set according to the number of simplified IGBP classes encountered in the sub-sampled training data. The cluster segments are learned completely unsupervised. The re-ordering of cluster labels is done via the Kuhn-Munkres algorithm \cite{Munkres_1957}, with the given low-resolution MODIS-derived labels of the sub-sampled train split serving as a reference. Clustering is done with the best-fitting of 10 $k$-means++ initializations \cite{Arthur_Vassilvitskii_2006}, each fitted for up to 300 iterations.
\item \emph{Random Forests (RF)}\newline supervised training on low-resolution MODIS-derived labels of the previously mentioned \emph{SEN12MS} subset. The model consists of an ensemble of $100$ trees, each with a maximum depth of $10$ nodes.
\end{itemize}\vspace{-8pt}
As can be seen from Tab.~\ref{tab:results_lowResHighRes}, shallow classifiers doing simple pixel-wise classification are not capable of reaching the baseline accuracy provided by the low-resolution labels and perform significantly worse than the deep learning models. Interestingly, for $k$-means, a fusion of Sentinel-1 and Sentinel-2 data deteriorates the result, which seems to be mainly caused by the classes \emph{Croplands} and \emph{Barren}.
Another fact to note is that the predicted maps displayed in Fig.~\ref{fig:pixelwise_segm} show more spatial details than the results achieved by the deep learning models, albeit at worse semantic accuracy. The existence of some spatial coherence in those pixel-based maps shows that the numerically observed misclassifications are of systematic nature. The Appendix provides supplementary results for weakly supervised learning of shallow classifiers trained directly on the target data, i.e. without spatial generalization.

\subsection{Data Fusion}
All machine learning models have used either the ten surface-related bands of Sentinel-2 as input, or have relied on a form of early data fusion by combining this Sentinel-2 input with the two polarimetric channels of Sentinel-1. As can be seen from Tab.~\ref{tab:results_lowResHighRes}, the fusion of Sentinel-1 and Sentinel-2 data only leads to slightly better results than what is achievable if only Sentinel-2 is used in the RF case. The fact that the fusion doesn't seem to help to improve the metrics achieved with the deep learning models should not be misunderstood: Since Tab.~\ref{tab:results_lowResHighRes} collects the best validation results, a fair comparison among the four deep semantic segmentation setups is not ensured. However, it will certainly be necessary to develop more sophisticated fusion procedures than simple band concatenation, e.g. with sub-networks that take the different data peculiarities into account as, e.g. proposed in \cite{Gawlikowski2020}. 

\section{Discussion}\label{sec:Discussion}
The baseline results presented in Section~\ref{sec:Baseline} show that mapping land cover on a global scale with models learned on inaccurate and inexact training labels remains an exciting challenge. 

In particular, it is interesting to note that three classes get consistently bad metrics throughout all classification methods (cf. Tab.~\ref{tab:results_lowResHighRes}): \emph{Shrublands}, \emph{Wetlands} and \emph{Barren}. As can be seen in Fig.~\ref{fig:class_occurence}, those three classes are the least frequent in the \emph{SEN12MS} dataset (except the understandably rare \emph{Snow/Ice} class). On the other hand, \emph{Wetlands} are massively over-represented in the \emph{DFC2020 validation} set. 

Figure~\ref{fig:transMat_LR_HR} shows that all three problematic classes seem to be significantly mislabeled in the low-resolution land cover maps -- while \emph{Wetlands} and \emph{Barren} areas used to label pixels actually containing water, \emph{Shrubland} pixels often are represented as \emph{Barren} in he high-resolution reference. 

Finally, it seems very promising that the \emph{Grassland} class is apparently not well represented by the low-resolution MODIS-derived labels, but can be well predicted by most models besides the unsupervised $k$-means. This indicates the potential of the topic addressed in this paper.

All in all, it becomes apparent that good models would have to solve two challenges: The transfer of spatial information from the training data sampled by the \emph{SEN12MS} dataset to the location of interest; and the transfer of land cover annotations from low resolution and noisy quality to high resolution and better quality. Standard methods without dedicated adaptations are apparently limited in their capabilities to do so.  

\section{Summary \& Conclusion}
In this paper, we have used the \emph{SEN12MS} dataset and the data provided in the frame of the IEEE-GRSS 2020 Data Fusion Contest to address the challenge of learning semantic segmentation models for global land cover mapping from inaccurate and inexact labels. While standard shallow and deep learning approaches were shown to already provide promising mapping capabilities, the results are not satisfying enough yet to consider off-the-shelf approaches for operational solutions. Therefore, we argue that specific models from the field of weakly supervised machine learning must be developed and expect that they will contribute greatly to a regular and fully automatic satellite-based monitoring of global land cover.

\section*{ACKNOWLEDGEMENTS}\label{ACKNOWLEDGEMENTS}
The authors would like to thank the Chairs of the IEEE-GRSS IADFC, N. Yokoya, R. H\"ansch and P. Ghamisi, for making the challenge of weakly supervised learning for global land cover mapping the topic of the 2020 IEEE-GRSS Data Fusion Contest; and for numerous fruitful discussions during the design of the contest.
\FloatBarrier

{
	\begin{spacing}{1.0}
		\normalsize
		\bibliography{isprs} 
	\end{spacing}
}

\vspace{-0.5cm}
\section*{APPENDIX: Weakly Supervised Learning Without Spatial Generalization}\label{APPENDIX}
\begin{figure}[!hb]
\centering
\begin{tabular}{rcc}
 \rotatebox{90}{$k$-means} &
 \includegraphics[width=0.38\linewidth]{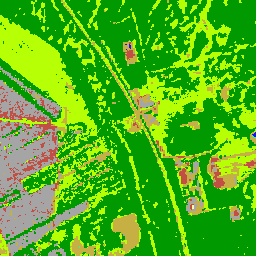}    &
 \includegraphics[width=0.38\linewidth]{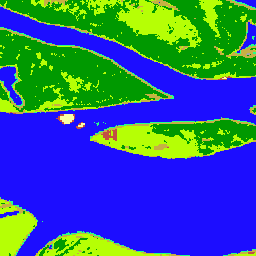}    \\
 \rotatebox{90}{RF} &
 \includegraphics[width=0.38\linewidth]{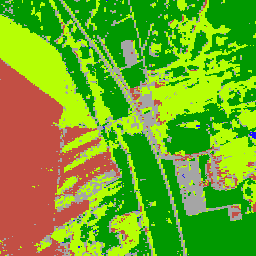}  & 
 \includegraphics[width=0.38\linewidth]{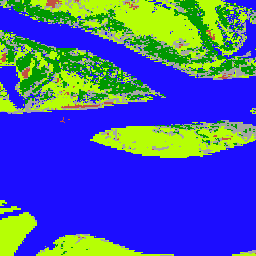}  \\
\end{tabular}
\caption{Qualitative results achieved with the shallow classifiers trained on the low-resolution labels of the \emph{DFC2020 validation} set.}
\label{fig:pixelwise_segm_overfitting}
\end{figure}

Of course, the question may arise whether the problem of weakly supervised learning for semantic segmentation of satellite images for land cover mapping could be simplified by aiming at less generic models. One way to create a less generic model is to forgo the desire to encode spatial generalization required by a globally applicable model and to train a scene-dependent model instead. To provide a sanity check, we have trained the shallow classifiers described in Section~\ref{sec:shallow} not on any data of the \emph{SEN12MS} dataset, but only on the low-resolution labels included in the \emph{DFC2020 validation} set. This follows the rationale that such labels are available for every location on the globe. The results are depicted in Tab.~\ref{tab:results_overfitting} and Fig.~\ref{fig:pixelwise_segm_overfitting}. It can be seen that these results exceed the quality of the results achieved for the scene-agnostic models, even though only shallow classifiers and only $986$ training samples were used. Apparently, transferring noisy, low-resolution labels into more accurate, high-resolution labels is a much simpler task than transferring land cover labels from one region of the globe to another region. 

 \begin{table}[h]
    \centering
    \footnotesize
    \begin{tabular}{L{1.3cm} R{1cm} R{1cm} R{1cm} R{1cm}}
     \toprule
     Class & $k$-means S2~only & $k$-means S1+S2& RF S2~only & RF S1+S2\\
       \cmidrule(r){1-1}  \cmidrule(lr){2-2} \cmidrule(lr){3-3} \cmidrule(lr){4-4} \cmidrule(l){5-5} 
Forest 		& 80.7\% 	& 93.3\% 	& 80.1\% 	& 80.1\%	\\
Shrubland 	& 0.3\% 	& 44.7\% 	& 0.9\% 	& 0.8\% 	\\
Savanna     & -- 		& -- 		& -- 		& --		\\
Grassland 	& 21.2\% 	& 49.8\% 	& 78.0\% 	& 78.2\%	\\
Wetlands    & 38.2\% 	& 1.3\% 	& 0.0\% 	& 0.0\% 	\\
Croplands 	& 33.4\% 	& 40.3\% 	& 80.7\% 	& 80.9\% 	\\
Urban 		& 38.8\% 	& 50.7\% 	& 91.8\% 	& 91.7\% 	\\
Snow/Ice 	& -- 		& -- 		& -- 		& --		\\
Barren 		& 0.4\% 	& 9.8\% 	& 0.0\% 	& 0.0\%		\\
Water 		& 73.1\% 	& 48.7\% 	& 99.9\% 	& 99.8\%	\\
\textbf{Average}&\textbf{35.8\%} & \textbf{42.3\%} & \textbf{54.0\%} & \textbf{54.1\%}\\
         \bottomrule
    \end{tabular}
    \caption{Quantitative results achieved on the \emph{DFC2020 validation} dataset for the shallow classifiers trained on the low-resolution labels of the \emph{DFC2020 validation} set.}
    \label{tab:results_overfitting}
\end{table}

\end{document}